\documentclass{article}
\usepackage{arxiv_template}
\BiblatexSplitbibDefernumbersWarningOff

\usepackage{amsmath,amsfonts,bm}

\def\eqref#1{equation~\ref{#1}}

\def\1{\bm{1}}

\DeclareMathAlphabet{\mathsfit}{\encodingdefault}{\sfdefault}{m}{sl}
\SetMathAlphabet{\mathsfit}{bold}{\encodingdefault}{\sfdefault}{bx}{n}

\usepackage{amssymb}
\usepackage{amsthm}
\usepackage{xurl}

\usepackage{booktabs}  
\usepackage{mdframed}
\usepackage{enumitem}
\setlist[enumerate]{nosep, topsep=0pt, itemsep=2pt, leftmargin=3em}
\usepackage{cleveref}

\newtheorem{theorem}{Theorem}[section]
\newtheorem{proposition}[theorem]{Proposition}

\newtheorem{corollary}[theorem]{Corollary}

\theoremstyle{definition}
\newtheorem{definition}[theorem]{Definition}

\theoremstyle{remark}

\title{The Linear Representation Hypothesis Needs a Group Action}

\author{
  \name{Louie Hong Yao}$^{1}$,
  \name{Yuhao Li}$^{2}$,
  \name{Shengchao Liu}$^{2}$\\[0.5em]
  \addr{$^{1}$Independent Researcher}\quad
  \addr{$^{2}$The Chinese University of Hong Kong}\\[0.35em]
  \textit{lhyao731@gmail.com}\quad
  \textit{\{yuhao.li, scliu\}@cuhk.edu.hk}
}

\begin{document}

\maketitle

\begin{abstract}
To make claims about representations that generalize beyond a particular trained model, we need to specify when two representations should count as equivalent. The Linear Representation Hypothesis is often discussed without making this equivalence explicit. Different notions of equivalence preserve different structures, so metrics, probes, and interventions that appear to study the same representation may in fact correspond to different hypotheses. We therefore argue that the Linear Representation Hypothesis is not one hypothesis but a family of claims distinguished by representation equivalence. We formalize this idea using group actions, specifying the representation object, the procedure that produces it, and the property ultimately asserted, while accounting for equivalences imposed by the model architecture. This framework clarifies how assumptions can change across metrics, reading points, and analysis stages, and we use it to audit common representation quantities and recent interpretability analyses.
\end{abstract}

\section{Introduction}
\label{sec:introduction}

The Linear Representation Hypothesis underlies much of modern interpretability, whether it is tested explicitly~\citep{park2024linear,marks2024geometry} or assumed when constructing linear probes~\citep{alain2016understanding,gurnee2024language,yao-etal-2026-rhetorical}, steering interventions~\citep{rimsky2024steering, liu2026cross,raval2026curveball}, learned dictionaries~\citep{bricken2023towards,templeton2024scaling}, and methods for comparing representations~\citep{kornblith2019similarity,williams2021generalized,domenichelli-etal-2026-linguistic}. Such studies usually aim to draw conclusions about representations rather than artifacts of a particular realization, which requires specifying when two realizations count as the same representation. Calling a representation ``linear'' does not answer this question: one must also specify which changes are merely changes of description.

Existing formulations rarely make this equivalence explicit. Different analyses implicitly choose different answers, so methods that appear to study the same representation may in fact be probing different objects and testing different hypotheses. The issue can arise even within a single study.

\citet{arditi2024refusal}, for example, estimate a difference-in-means vector and use the same array in two ways. It is added to activations without normalization, where it acts as a displacement in $V$ and its magnitude affects the intervention, and separately normalized before its span is projected out, where only its projective direction in $\mathbb{P}(V)$ matters. Both interventions are effective and are reported as evidence for a single refusal direction, even though the two uses place that direction in different mathematical spaces and require different notions of equivalence.

Neither operation is invalid, since each is well defined under its assumed structure. The problem is that these assumptions are often unstated and may change between estimation, comparison, and intervention. As a result, different stages of an analysis can silently refer to different notions of the representation, while the final conclusion is stated as though they referred to the same object.

We propose that a representation claim be specified by the space $M$ in which its object lives together with the action of an equivalence group $G$, the procedure $F$ that produces the object, and the predicate $P$ ultimately asserted of it. The procedure must transform equivariantly under changes of representation, while the predicate must remain invariant. The choice of $G$ is also constrained by function-preserving reparameterizations of the architecture. If a function-preserving reparameterization changes the value of a quantity computed from an internal representation, that quantity is a property of the parameterization rather than of the model.

Using this specification, we show that the Linear Representation Hypothesis is not one hypothesis but a family of claims. Displacement, decoding, superposition, and subspace formulations place features in different mathematical objects and admit different transformation laws. The analogous distinction applies to the procedures used to estimate these objects: two procedures may produce objects of the same type while respecting different equivalences. We further show that admissible symmetry depends on where a representation is read and that the symmetry of a multi-stage analysis must be checked for the composition as a whole.

Our position is that this specification is part of stating a representation claim rather than metadata attached to it. Making it explicit may narrow some conclusions, but it makes clear when two analyses are genuinely studying the same representation 
hypothesis.

The paper proceeds as follows. Sections~2--3 motivate and formalize representation equivalence. Section~4 develops its architectural and compositional consequences. Sections~5--6 audit common quantities and recent interpretability analyses. Section~7 extends the argument to the Platonic Representation Hypothesis, and Section~8 states the reporting principle.


\section{The Linear Representation Hypothesis Is Not One Hypothesis}
\label{sec:not-one}

The Linear Representation Hypothesis is not a single claim. Different formulations place features in different mathematical objects and admit different transformation laws under which those objects remain meaningful. We first review some common formulations below.\footnote{Notation and basic mathematical preliminaries are summarized in Appendix~\ref{app:notation}.}

A classical formulation treats a feature or relation as a \emph{displacement}. In word embeddings, some lexical relations were found to satisfy approximately constant offsets, $h_b-h_a\approx v$ \citep{mikolov2013linguistic}. Modern methods such as difference-in-means \citep{marks2024geometry} and contrastive activation addition \citep{rimsky2024steering} use the same basic object. Under an affine transformation $h'=Ah+t$, the displacement transforms as $v'=Av$, since the translation cancels. The coordinates of the vector change, but the constant-offset relation does not. This formulation therefore places the feature in $V$, with the primal affine action $v \mapsto Av$.

Linear decodability provides a different formulation. If a property can be recovered from an activation $h\in V$ by a linear functional $w\in V^*$ through $w^\top h$ \citep{alain2016understanding}, then an invertible transformation $h'=Ah$ is accompanied by the dual action $w'=A^{-\top}w$, which preserves the readout. With a bias term, $w^\top h+b$, this extends to affine transformations by taking $b'=b-w'^\top t$. Thus even within probing, the admissible group depends on whether the probe includes a bias term. More importantly, the object has changed: a probe lives in $V^*$ rather than $V$. Although probe weights and steering vectors are both stored as $d$-dimensional arrays, they obey different transformation laws and cannot be canonically identified without additional structure. Related work similarly distinguishes measurement from intervention geometry and linear representation from linear decodability \citep{park2024linear, park2025geometry, garg2026many}.

Superposition gives another meaning to linear representation. In this formulation, an activation is written as
$h=b+\sum_i x_iW_i$,
where $W_i$ denotes a feature direction and $x_i$ its coefficient \citep{elhage2022toy}. This additive view later motivated dictionary-learning approaches and sparse autoencoders \citep{bricken2023towards,templeton2024scaling}. As a claim about the activation space, the decomposition is affine-covariant: taking $b\mapsto Ab+t$ and $W_i\mapsto AW_i$ reproduces the transformed activation with the coefficients unchanged. It also has a distinct ambiguity in its latent coordinates, reduced but generally not eliminated by sparsity or other constraints (Section~\ref{sec:audit}). This latent non-identifiability is distinct from equivalence under changes of activation coordinates.

These examples show that ``linear representation'' does not determine a unique object or transformation law. Moreover, even for a fixed object space, different procedures can impose different symmetry requirements. We separate these choices before formalizing them in \Cref{sec:specifying a member}.

The object itself must be specified. Linearity also need not imply that a feature is one-dimensional \citep{olah2024linear}. In our framework, a one-dimensional feature naturally lives in $\mathbb{P}(V)$, while a $k$-dimensional linear feature lives in $\mathrm{Gr}(k,V)$. Circular representations of days and months provide an example in which a two-dimensional structure mediates computation and no single direction suffices \citep{engels2025not}. This distinction is compatible with linearity because the object space and admissible transformation group are separate parts of the claim: an invertible linear map preserves both lines and \(k\)-dimensional subspaces.

The object alone also does not determine the symmetry of the claim. A difference-in-means direction and a leading principal direction can both lie in $\mathbb{P}(V)$, but the procedures that produce them have different symmetry requirements. Difference-in-means requires no inner product, whereas PCA requires one to order directions by explained variance. Thus two procedures may return objects in the same space while being equivariant under different groups.

This distinction has an immediate consequence for empirical evidence. None of the formulations above intrinsically requires an inner product. Yet common evidence for them does: cosines and spectral quantities require similarity geometry, while absolute norms and intervention magnitudes require a fixed scale. The metric therefore enters through the evidence used to support the hypothesis rather than through the hypothesis itself.

This perspective differs from work that fixes a nuisance group to define representation similarity. In the generalized shape metrics of \citet{williams2021generalized}, for example, quotienting by a chosen group yields a metric on representation space. Our position differs. The group is presupposed by the claim rather than chosen to make a measurement well behaved, so it is a property of the hypothesis. And it also constrained by the architectural floor and by the analysis pipeline (Sections~\ref{subsec:architectural-floor} and~\ref{subsec:composition}).


\section{Specifying a Member of the Family}
\label{sec:specifying a member}

\Cref{sec:not-one} showed that formulations of the Linear Representation Hypothesis must specify not only what object represents a feature, but also how that object transforms and how it is obtained. We now formalize these choices as a specification of a representation claim.

\subsection{Representation Equivalence}
\label{subsec:same_representation}

Let $\mathcal{X}$ be an input domain, $V$ a finite-dimensional vector space, and $\mathcal{S}=\{\phi:\mathcal{X}\to V\}$ the space of representations.
Suppose a group $G$ acts on $\mathcal{S}$. In the settings considered below, this action is pointwise on the representation values in $V$: for each $g\in G$, there is a transformation $T_g:V\to V$ such that
\begin{equation}
    (g\cdot\phi)(x)=T_g(\phi(x)).
\label{eq:representation-action}
\end{equation}
This action specifies which transformations count as changes of description rather than changes in the underlying representation.

\begin{definition}[Representation equivalence]
\label{def:equivalence}
Two representations $\phi_1,\phi_2\in\mathcal{S}$ are equivalent under $G$, written $\phi_1\sim_G\phi_2$, if $\phi_2=g\cdot\phi_1$ for some $g\in G$. The equivalence class of $\phi$ is its orbit
$[\phi]_G=\{g\cdot\phi:g\in G\}$.
\end{definition}

The choice of equivalence group determines the strength of the representation claim. A larger group leaves fewer quantities invariant, while a smaller group risks promoting coordinate-specific properties to properties of the representation. The equivalence relation is therefore part of the substantive claim, and it is not always freely chosen. A function-preserving reparameterization may alter the coordinates of an internal representation, so a claim about the model rather than one parameterization must remain valid across such realizations. The admissible group must therefore contain the transformations the architecture already realizes, a constraint made precise in Section~\ref{sec:consequences}.

\subsection{Common Choices of Transformation Group}
\label{subsec:chain}

We now specialize the pointwise transformations in
\Cref{eq:representation-action} to affine maps
$T_g(h)=Ah+t$, where $A$ and $t$ may depend on $g$.
Within this class, three common transformation groups form the hierarchy
\begin{equation}
    G_{\mathrm{iso}}
    \subset
    G_{\mathrm{sim}}
    \subset
    G_{\mathrm{aff}}.
\end{equation}
The affine group $G_{\mathrm{aff}}$ allows arbitrary invertible $A$ and translations $t$. The similarity group $G_{\mathrm{sim}}$ restricts the linear part to $A=sQ$, where $s>0$ and $Q$ is orthogonal. And the isometry group $G_{\mathrm{iso}}$ further requires $s=1$.

These groups preserve progressively stronger structures. Affine transformations preserve affine relations, collinearity, subspace dimension, and intersection structure, but not angles or lengths. Similarities additionally preserve angles, orthogonality, and cosine similarity. Isometries further preserve norms, distances, and fixing absolute scale. A larger equivalence group therefore imposes stronger invariance requirements, while allowing fewer quantities to be attributed to the representation.

\subsection{The Object and the Procedure}
\label{sec:action-object}

The equivalence group does not fully specify a representation claim. An analysis must also specify what mathematical object is extracted from the representation and how that object is constructed.
In practice, the construction uses finitely many sampled activations
$\{\phi(x_i)\}_{i=1}^n$. Under the pointwise actions of \Cref{subsec:chain}, these samples transform with the same coordinate change.

Let $M$ denote the space of mathematical objects, equipped with an action of $G$, and let $F:\mathcal{S}\to M$ denote the procedure that constructs the object. A claim is a predicate $P$ on $M$, so the resulting statement about $\phi$ is $P(F(\phi))$. For the statement to be independent of the chosen representation coordinates, the procedure and predicate must satisfy
\begin{equation}
F(g\cdot\phi)=g\cdot F(\phi),
\qquad
P(g\cdot m)=P(m).
\label{eq:equivariance}
\end{equation}
It then follows that
$P(F(g\cdot\phi))
= P(g\cdot F(\phi)) = P(F(\phi))$.
Notice that the condition \Cref{eq:equivariance} is stronger than invariance of the composite predicate $P \circ F$. We impose this factorization because the intermediate object is itself part of the representation claim and may be used in subsequent analyses. Its transformation law is therefore substantive: specifying $M$ requires specifying not only what kind of object it contains, but also how $G$ acts on that object.

This distinction already matters for two of the most common objects in representation analysis. A displacement is naturally an element of $V$ and transforms as $v\mapsto Av$, whereas a linear probe is naturally an element of $V^*$ and transforms as $w\mapsto A^{-\top}w$. 
Although both are vectors in the implementation, 
they belong to different $G$-spaces. The following proposition makes precise what additional structure is required to identify them.

\begin{proposition}[Primal and dual objects]
\label{prop:primal-dual}
Let $\dim V \ge 2$ and let $\mathrm{GL}(V)$ act on $V$ by
$v \mapsto Av$ and on $V^{*}$ by $w \mapsto A^{-\top}w$.
Then:
\begin{enumerate}[label=\roman*.]
\item There is no nonzero $\mathrm{GL}(V)$-equivariant map $V^*\rightarrow V$.
\item An inner product $g$ induces an equivariant map $\sharp_g : V^{*} \to V$ under $G_{\mathrm{iso}}$.
\item The induced map $[w] \mapsto [\sharp_g w]$ on projective spaces is equivariant under $G_{\mathrm{sim}}$.
\end{enumerate}
\end{proposition}

A metric thus supplies an identification between $V^*$ and $V$ only under isometries, while passing to projective directions removes sensitivity to uniform scale and enlarges the symmetry to similarities.
Regularized probe fitting provides another example: the $\ell_2$ penalty breaks equivariance under general $\mathrm{GL}(V)$ transformations, leaving only isometric equivariance.

\subsection{The Specification}
\label{sec:specification}

The preceding components can now be collected into a complete specification:
\begin{equation*}
\text{\bfseries Complete specification requires }G,\ M,\ F,\ \text{and }P.
\end{equation*}

\begin{definition}[Representation claim]
\label{def:claim}
Let group $G$ act on space $\mathcal{S}$. A representation claim consists of a $G$-space $M$, a map $F:\mathcal{S}\rightarrow M$, and a predicate $P$ on $M$. The claim about $\phi\in\mathcal{S}$ is $P(F(\phi))$.
\end{definition}

The $G$-space specifies both the mathematical object and how it transforms. The map $F$ is separate because the same object space may be reached by procedures with different symmetry properties.

\begin{definition}[Admissibility]
\label{def:admissible}
A representation claim $(G, M,F,P)$ is admissible under $G$ if
\begin{enumerate} 
\item[(A1)] $F$ is equivariant and $P$ is invariant under the declared actions, as in \Cref{eq:equivariance}.
\item[(A2)] The action of $G$ on $\mathcal{S}$ contains the architecture-induced equivalences described in \Cref{sec:consequences}.
\end{enumerate}
\end{definition}

Condition (A1) is internal to the analysis: it requires the method and conclusion to be well defined on the declared equivalence classes rather than on a selected coordinate realization. Condition (A2) supplies an external floor. If the architecture realizes a function-preserving transformation, a claim about the model cannot distinguish representations related by it. An analysis may therefore choose a larger group, but not a smaller one than the architecture permits.

\begin{definition}[Comparability]
\label{def:comparable}
Two admissible claims are directly comparable if they use the same representation equivalence and the same $G$-space. Claims on different $G$-spaces require an explicit equivariant map relating those spaces and are otherwise not directly comparable.
\end{definition}

Non-comparable claims may both be correct without supporting one another: equal numerical shape is insufficient unless the objects belong to spaces carrying compatible actions. When the object space is shared and $G_1\subseteq G_2$ with the $G_1$ action obtained by restriction, (A1) under $G_2$ implies (A1) under $G_1$, although (A2) must be rechecked. Restricting the group can therefore make additional predicates well defined, as when an object estimated under a larger group is later reported through angles or norms requiring a smaller one.


\section{Two Consequences of the Specification}
\label{sec:consequences}

The specification has two immediate consequences: an external constraint imposed by the architecture (A2), and an internal constraint arising from composition (A1).

\subsection{The Architectural Floor}
\label{subsec:architectural-floor}

The equivalence group is not always freely chosen. 
Some transformations arise from the architecture itself at the parameter level and constrain which equivalence groups are admissible.
Let $\Theta$ be the parameter space, let $f_\theta$ denote the function computed by parameters $\theta$, and define the group of function-preserving parameter transformations as
\begin{equation}
    \Gamma=\{\gamma:\Theta\to\Theta \mid \gamma \text{ invertible and } f_{\gamma\cdot\theta}(x)=f_\theta(x)\text{ for all }\theta,x\}.
\end{equation}

At a fixed reading point, a parameter symmetry induces a representation action when the transformation descends to the activations.
Writing $\phi^\theta:\mathcal{X}\to V$ for the representation at a fixed reading point, such an action exists when $\phi^{\gamma\cdot\theta}(x)=\rho(\gamma)\phi^\theta(x)$ for all $x\in\mathcal{X}$. The image of $\rho$ is the architectural symmetry group at that reading point.

\begin{proposition}[Architectural invariance]
\label{prop:architectural-floor}
Suppose $\gamma\in\Gamma$ induces $\rho(\gamma)$ at a reading point, and let $(G,M,F,P)$ be a representation claim satisfying (A1). If $P(F(\phi^{\gamma\cdot\theta})\neq P(F(\phi^\theta))$ for some $\theta\in\Theta$, then the claim is not a property of the model function: it takes different values on parameter settings that compute the same function.
\end{proposition}
Thus condition (A2) requires $G \supseteq \mathrm{Im}\rho$ at the reading point.
\begin{corollary}[Reading-point dependence]
\label{cor:reading-point}
The same architecture can induce different symmetries at different reading points.
In dot-product attention, $W_Q\mapsto \Lambda W_Q$ and $W_K\mapsto \Lambda^{-\top}W_K$ preserve the model function for any $\Lambda\in\mathrm{GL}(d_{\mathrm{head}})$. At the residual stream this transformation acts trivially, so it excludes no angular claim there. At a query or key site, however, $\operatorname{Im}\rho$ contains $\mathrm{GL}(d_{\mathrm{head}})$, which admits no nonzero invariant bilinear form: taking $\Lambda = cI$ gives $b(cv,cw) = c^{2}b(v,w) = b(v,w)$ for all $c>0$, forcing $b = 0$. Angular and metric predicates defined from a fixed inner product are therefore not invariant at those sites.
\end{corollary}

Architecture-induced symmetries of internal representations have been studied previously \citep{godfrey2022symmetries}, including joint query--key rotations in transformers \citep{zhang2025beyond}. 
With RoPE, the induced symmetry is reduced but remains generally anisotropic, so the same conclusion holds.
The derivation is given in Appendix~\ref{app:arch}.

Equal-dimensional representations at different reading points may carry different $G$-actions, so transporting a direction between them requires an explicit equivariant map in the sense of Definition~\ref{def:comparable}, rather than identification by shared coordinates.

\subsection{Composition of Analysis Stages}
\label{subsec:composition}

Representation analyses are usually pipelines rather than single maps. Therefore, admissibility must be established for the composite procedure rather than inferred from individual stages.

\begin{proposition}[Composition]
\label{prop:composition}
Let $F=F_n\circ\cdots\circ F_1$ with $F_i:M_{i-1}\to M_i$. If $G$ acts on every $M_i$, each $F_i$ is equivariant under the corresponding actions, and $P$ is invariant on $M_n$, then $(G, M_n,F,P)$ satisfies (A1) under $G$.
\end{proposition}

If a stage is not equivariant, admissibility must instead be established for the composite. Its symmetry is not generally obtained by intersecting groups assigned to the stages in isolation, because a stage may change the object space and hence the action seen by the next stage. For example, under $h\mapsto Ah+t$, centered activations transform as $h-\bar h\mapsto A(h-\bar h)$, so cosine after centering can be invariant to translations that would change cosine on the uncentered activations.

When all stages carry compatible actions of a common group, the pipeline is limited by its most restrictive stage. A direction may be estimated under $G_{\mathrm{aff}}$, compared by cosine under $G_{\mathrm{sim}}$, and calibrated by a norm under $G_{\mathrm{iso}}$, so the resulting claim is guaranteed admissible only under $G_{\mathrm{iso}}$ unless stronger invariance of the composite is established.


\section{A Symmetry Audit of Common Representation Quantities}
\label{sec:audit}
Table~\ref{tab:audit} records, for quantities in common use, the largest group within the hierarchy considered here under which each is well defined; using a smaller group narrows the claim it can support.


\begin{table}[ht]
\centering
\small
\renewcommand{\arraystretch}{0.95}
\caption{Symmetry audit of common representation quantities. Rows are grouped by the largest group under which the quantity is preserved, generically in the spectral parameters. The final column names the structure that fixes the restriction.}
\label{tab:audit}
\begin{tabular}{p{0.35\linewidth} p{0.28\linewidth} p{0.28\linewidth}}
\toprule
Quantity & Computed from & What fixes the group \\
\midrule
\multicolumn{3}{l}{\emph{Preserved under $G_{\mathrm{aff}}$}} \\
Exact rank & Centered activations & Linear dependence only \\
Span & Centered activations & Linear dependence only \\
Containment, intersection dim. & Subspace pair & Incidence structure \\
CCA & Representation pair & Centered linear relations \\
\midrule
\multicolumn{3}{l}{\emph{Preserved under $G_{\mathrm{sim}}$}} \\
Leading principal subspace & Covariance form & Variance ordering \\
Relative-threshold rank & Covariance spectrum & Spectral ratios \\
Effective rank, participation ratio & Covariance spectrum & Spectral ratios \\
Intrinsic dimension & Neighbour distances & Distance ratios \\
Cosine, angle, orthogonality & Direction pair & Inner product up to scale \\
Principal angles & Subspace pair & Inner product up to scale \\
Grassmann distances & Subspace pair & Inner product up to scale \\
Linear CKA & Representation pair & Normalized Gram geometry \\
Nearest-neighbour agreement & Representation pair & Distance ordering \\
\midrule
\multicolumn{3}{l}{\emph{Preserved under $G_{\mathrm{iso}}$}} \\
Norm, distance & Activations & Fixed Euclidean scale \\
Intervention magnitude & Displacement in $V$ & Norm and scale \\
Absolute-threshold rank & Covariance spectrum & Absolute threshold \\
Procrustes distance & Representation pair & Euclidean alignment loss \\
Reconstruction loss & Residual in $V$ & Euclidean residual norm \\
\midrule
\multicolumn{3}{l}{\emph{Preserved under signed permutations of the latent coordinates}} \\
Coordinatewise sparsity penalty & Latent codes & Coordinatewise axes \\
\bottomrule
\end{tabular}
\end{table}

\noindent\textbf{Dictionary learning acts on two spaces.}
Under $h\mapsto Ah+t$, taking $D\mapsto AD$, $W_{\mathrm{enc}}\mapsto W_{\mathrm{enc}}A^{-1}$, and transforming the encoder and decoder biases to absorb $t$ leaves every encoder preactivation, and hence the latent codes, unchanged. The reconstruction residual transforms as $h-\hat h\mapsto A(h-\hat h)$, so the Euclidean loss is preserved for all residuals only when $A^\top A=I$. The restriction to $G_{\mathrm{iso}}$ on the activation side therefore comes from the reconstruction loss \citep{bricken2023towards,templeton2024scaling}. On the latent side, $W\mapsto WB$ and $x\mapsto B^{-1}x$ leave the reconstruction unchanged, while a coordinatewise sparsity penalty reduces this mixing symmetry to signed permutations. These are distinct restrictions on distinct spaces.


\section{Current Practice Leaves the Specification Implicit}
\label{sec:practice}

The preceding sections make the specification explicit. We now examine what happens when its components remain implicit. The recurring problem is not that strong structural assumptions are necessarily unwarranted, but that objects, actions, and procedures are identified or changed without recording the corresponding change in the claim.

\subsection{Identification by Storage Format}
\label{subsec:storage}

A linear probe produces a coefficient vector in $V^*$, while a steering method such as difference-in-means produces a displacement in $V$. Because both are stored as length-$d$ arrays, they are routinely treated as the same kind of direction. Probe weights are compared to steering vectors by cosine, used as intervention directions, or combined with vectors of other provenance. For example, \citet{bhalla2024towards} use both steering vectors and linear-probe weights as additive intervention directions and compare their directions by cosine similarity. The same identification appears across reading points. A direction estimated at one layer is often transported to another by the identity map because both residual streams have dimension $d$, even though they need not carry the same group action.

Proposition~\ref{prop:primal-dual} separates two operations that this practice conflates. Once a metric is fixed, the induced map from $V^*$ to $V$ descends to projective spaces equivariantly under $G_{\mathrm{sim}}$. A cosine comparison between a probe direction and a steering direction can therefore be meaningful under similarity geometry. Reusing the probe coefficients themselves as an additive displacement is stronger. The vector-level identification is equivariant only under $G_{\mathrm{iso}}$.

\subsection{Normalization and Calibration}
\label{subsec:normalization}

Hidden group choices also enter through normalization and calibration. Contrastive Activation Addition estimates a difference-in-means vector and applies it as a scaled translation \citep{rimsky2024steering}. The reported procedure normalises vector magnitudes across behaviours but not across layers, where residual-stream norms grow over the forward pass. These choices assign meaning to length and therefore introduce Euclidean scale into an otherwise affine-covariant construction. Activation Addition adopts the opposite convention \citep{turner2024steering}: its activation-difference vector is left unnormalised, so the displacement produced by a coefficient depends on the norm supplied by the sampled activations. The same numerical coefficient therefore does not denote the same displacement under the two conventions. \citet{wollschlager2025the} make the required calibration explicit by scaling an optimised refusal direction to match the norm of a difference-in-means direction, thereby stipulating a common Euclidean magnitude.

Because directional ablation depends only on the normalised direction, \citet{wollschlager2025the} sample unit directions within refusal cones directly rather than normalising arbitrary convex combinations, which would bias the induced distribution over directions. In our terminology, the sampling procedure is adapted to the projective object actually consumed by the intervention.

\subsection{Unnamed Reading Points}
\label{subsec:reading-points}

The reading point is itself part of the $G$-space. Corollary~\ref{cor:reading-point} shows that the same function-preserving parameter transformation can act trivially at the residual stream and as a nontrivial general linear transformation at a query or key site.

KeyDiff provides a direct case \citep{park2026keydiff}. The method evicts KV-cache entries using pairwise cosine similarity among keys within a head, and its supporting analyses report key and query cosines, their $L^{2}$ norms, PCA of key caches, and $\log\det(KK^{\top})$, all in per-head query or key space. The empirical performance of the eviction rule is not in question here. What is at issue is the status of the geometric quantities offered in explanation of it. In ordinary dot-product attention, the functionally relevant quantity is the bilinear pairing $q^\top k$, equivalently $W_Q^\top W_K$ at the parameter level \citep{elhage2021mathematical}; with RoPE, it is the family $W_Q^\top R(\tau)W_K$ indexed by relative position. By Corollary~\ref{cor:reading-point} and Appendix~\ref{app:arch}, this gauge is anisotropic even under RoPE, so none of these quantities is invariant. The quantities used to explain the computation are therefore not invariants of the computation they explain.

Related analyses of head-space quantities inherit the same dependence on chosen head coordinates. Cross-Gram singular values between head projections are not invariant under general invertible changes of basis, although the underlying residual-stream subspaces are, unless the spanning matrices are first orthonormalised \citep{fokouuavs2026multi}. Likewise, cosine comparisons of head-space singular vectors and their calibration against a fixed Euclidean reference distribution depend on the chosen head-space geometry \citep{ge2026prism}. Appendix~\ref{app:head-space} gives corresponding transformations.

By contrast, \citet{yamagiwa2026measuring} compare column spaces of transposed projection matrices, which are subspaces of the residual stream. Under $W_Q\mapsto \Lambda W_Q$, for example, $W_Q^\top\mapsto W_Q^\top \Lambda^\top$ leaves its column span unchanged. In our framework, the comparison targets an object preserved by the parameter symmetry rather than geometry internal to the gauge-dependent head coordinates.

Gauge-related parameter settings compute the same model function but can cause a cosine-based key-space eviction rule to select different tokens. Naming the reading point is therefore necessary to determine whether a reported geometric property belongs to the model, to an explicitly chosen geometry, or only to the sampled parameterization.

\subsection{Inconsistency Within a Single Analysis}
\label{subsec:within-analysis}

The clearest cases arise when the specification changes within one pipeline. Proposition~\ref{prop:composition} requires the conclusion to respect the actions introduced throughout the analysis, not merely the symmetry of its initial estimator. Three patterns recur.

\noindent\textbf{One estimate, different $G$-spaces.}
\citet{arditi2024refusal} use one difference-in-means estimate both as an unnormalised displacement in $V$ and as a projective direction in $\mathbb{P}(V)$ (Section~\ref{sec:introduction}). The estimate changes $G$-space between uses, and the combined conclusion is stronger than either use establishes.

\noindent\textbf{Change of object mid-analysis.}
A probe coefficient belongs to $V^*$, while an additive intervention consumes an element of $V$, so moving from probing to intervention requires specifying how the two spaces are related. \citet{li2023inferencetime} compare probe-weight, mass-mean-shift, and CCS directions as alternatives for intervention after using probes to identify relevant attention heads. Their intervention partially makes the required calibration explicit: the chosen direction is normalised, and the added displacement is scaled by the empirical standard deviation of activations along that direction. Under a similarity $A=sQ$, a normalised probe direction transforms as $Q\hat w$ while the projected standard deviation transforms as $s\sigma$, so the calibrated displacement $\sigma\hat w$ transforms covariantly as $A(\sigma\hat w)$. The complete construction is therefore compatible with $G_{\mathrm{sim}}$, while identifying a probe direction in $V^*$ with an intervention direction in $V$ still depends on the chosen similarity geometry.

\noindent\textbf{Preprocessing that restricts the group.}
SVCCA provides the failure at the level of procedures \citep{raghu2017svcca}. CCA alone is affine-invariant on centered representations, but SVCCA first performs singular-value truncation, a spectral operation not preserved by general invertible linear transformations. The CCA stage cannot restore the affine invariance lost in preprocessing.

In each case the problem is not the presence of a strong assumption. It is that the assumption enters at one stage while the conclusion is stated as though it inherited the weaker assumptions of another.

\section{The Same Omission Beyond Linear Representation}
\label{sec:beyond}

The same specification issue appears in the Platonic Representation Hypothesis \citep{huh2024platonic}, although the hypothesis itself is unrelated to linear representation. Representations of models with different widths need not lie in a common representation space, so cross-model comparison proceeds through structures induced on shared inputs, such as neighbourhoods or similarity matrices. Because two representations may induce the same structure on one input distribution and differ elsewhere, the resulting claim is also indexed by the distribution on which the comparison is made. These dependencies are typically left implicit.

Convergence then amounts to saying that differently realised representations belong to a common equivalence class. The object, predicate, and comparison procedure are supplied by the hypothesis, while the transformation group remains open, and it is this remaining component on which the truth value turns. On a finite sample, sufficiently permissive linear equivalence can become degenerate. \citet{kornblith2019similarity} show that when the representation width is at least the number of sampled inputs and both activation matrices have full row rank, a similarity measure invariant to arbitrary invertible linear transformations cannot distinguish them. Finite-sample convergence under such an equivalence can therefore become near-trivial in this regime. Under an isometric interpretation, by contrast, convergence would require agreement in absolute metric structure that models with different widths, tokenizers, and objectives generally cannot satisfy. Existing evaluations instead operate closer to $G_{\mathrm{sim}}$, with the effective equivalence determined by the alignment measures in use rather than specified independently of them.

This places the Platonic Representation Hypothesis directly inside the audit of Section~\ref{sec:audit}. Nearest-neighbour agreement, kernel-based alignment, and angular quantities inherit the same geometric commitments already catalogued there. The truth of the Platonic Representation Hypothesis cannot be assessed until the object being compared and the equivalence under which convergence is asserted have been specified.

\section{The Specification Principle}
\label{sec:principle}

One clarification is needed before stating the principle. Condition (A1) can always be satisfied by shrinking $G$ or weakening $P$, so admissibility alone carries no content. What carries content is maximality: the strength of a claim is the largest group under which it survives. Reporting a predicate under an unnecessarily small group is not an error of validity but a loss of content. This is why Table~\ref{tab:audit} is organized by the largest preserving group rather than by any group under which a quantity happens to be defined.

We therefore propose a simple reporting principle. A representation study should state the $G$-space in which its object lives, the procedure $F$ that produces the object and why it is equivariant under the declared action, and the predicate $P$ ultimately asserted of it. It should also name the reading point at which the representation is taken and check the symmetry of the complete analysis when several stages are composed. These requirements need not be presented in group-theoretic notation, but the underlying choices must be recoverable from the statement of the claim rather than reconstructed from the implementation.

This requirement has a real cost. Some conclusions become narrower once their geometric assumptions are made explicit, some comparisons require an additional map between spaces, and some quantities must be attributed to a representation together with a chosen metric or parameterization rather than to the representation alone. One might instead declare $G_{\mathrm{iso}}$ throughout, but Condition (A2) forbids this when the architecture realizes transformations outside $G_{\mathrm{iso}}$.

A second objection is that training returns one specific $\theta$, so the geometry of that $\theta$ is a fact about the artifact one actually possesses, whatever the orbit contains. This is correct, and the framework does not forbid it. It fixes what such a statement is about. A property that varies across $\operatorname{Im}\rho$ is a property of $\theta$ together with the procedure that produced it, not of $f_{\theta}$, and asserting it therefore incurs an empirical obligation that is rarely discharged: the reported geometry must be shown stable across seeds and training runs before it can be attributed to anything more general than the run in hand.

A third objection is that the proposal merely relabels assumptions already present in existing analyses. \citet{park2024linear,park2025geometry}, for example, make the geometry relating measurement and intervention explicit by specifying an inner product under which causally separable concepts are orthogonal. \citet{golechha2025intricacies} argue that near-orthogonality can arise generically in high dimensions and be induced by whitening, so the resulting geometry is not uniquely diagnostic of learned conceptual structure. Making the geometry explicit therefore does more than relabel an assumption: it turns the disagreement into a precise question about which inner product supports the orthogonality predicate.


\section{Conclusion}
\label{sec:conclusion}
The Linear Representation Hypothesis is not one hypothesis but a family, and its members are distinguished by which representations they treat as the same. Stating a member requires a $G$-space, an equivariant procedure producing the object, and a predicate asserted of it, subject to the floor the architecture already fixes. Once these are recorded, two studies can be checked for whether they assert the same claim without reconstructing both analyses. Nothing in the requirement depends on linearity, and the same omission appears wherever a claim about representations is made at all.

\section*{Acknowledgment}
L.H.Y. thanks Tianyu Jiang for helpful discussions.

\clearpage
\renewcommand*{\bibfont}{\small}
\printbibliography

\clearpage
\appendix

\section{Notation and Mathematical Preliminaries}
\label{app:notation}

\subsection{Groups and Symmetries}

A symmetry is a transformation that preserves a specified structure.
Mathematically, a collection of compatible transformations is described by a group. The choice of group determines which transformations are regarded as equivalent descriptions of the same object.

A group $G$ act on a set $\mathcal{X}$, if each element $g\in G$ assigns a transformation
\begin{equation}
    x \mapsto g \cdot x
\end{equation}
such that the identity transformation acts trivially and composition
of transformations follows the group operation. The orbit of an element $x$ under $G$ is
\begin{equation}
    [x]_G=\{g\cdot x:g\in G\}.
\end{equation}
The orbit defines the equivalence relation induced by the group action: two elements are equivalent if they belong to the same orbit.

\subsection{Linear Spaces}

Let $V$ be a finite-dimensional real vector space, and the dual space of $V$ is the vector space of linear functionals
\begin{equation}
    V^*=\{w:V\rightarrow \mathbb{R}\mid w \text{ is linear}\}.
\end{equation}
A vector $v \in V$ and a dual vector $w \in V^*$ are related through the natural pairing 
\begin{equation}
    \langle w,v \rangle = w(v).
\end{equation}
Although both $V$ and $V^*$ are finite-dimensional vector spaces with the same dimension, they are distinct spaces without additional structure such as an inner product. 

An affine space is a vector space without a fixed origin, in which the difference between points forms vector, and the addition of a point and a vector results in a new point. Given a vector space $V$, an affine transformation takes the form
\begin{equation}
    h \mapsto Ah + t,
\end{equation}
where $h$ and $t$ are vectors in $V$, and $A$ is an invertible linear transformation on $V$. Under an affine transformation, displacements transform as
\begin{equation}
    (h_2-h_1) \mapsto A(h_2-h_1),
\end{equation}
so affine relations are preserved while lengths and angles
are generally not.

The projective space of $V$ is the space of one-dimensional linear
subspaces of $V$
\begin{equation}
    \mathbb{P}(V) = \{\mathrm{span}(v): v\in V, v\neq0\}.
\end{equation}
Equivalently, projective space identifies vectors that differ only by
a nonzero scalar multiplication. Therefore, quantities defined on
$\mathbb P(V)$ describe directions rather than vector magnitudes.

The Grassmannian $\mathrm{Gr}(k, V)$ denotes the space of all $k$-dimensional linear subspaces of $V$
\begin{equation}
    \mathrm{Gr}(k, V) = \{ U \subseteq V: \mathrm{dim} (U) = k \}.
\end{equation}

The projective space $\mathbb P(V)$ is the special case $\mathrm{Gr}(1, V)$, where an element corresponds to a one-dimensional feature. More generally, a multidimensional linear feature is
naturally represented as an element of $\mathrm{Gr}(k,V)$.

For a vector space $V$, the general linear group is
\begin{equation}
    \mathrm{GL}(V)=\{A: V \rightarrow V \mid A \text{ is linear and invertible}\}.
\end{equation}
It acts naturally on vectors by $v \mapsto Av$. The induced action on the dual space is $w \mapsto A^{-\top} w$ which preserves the vector-dual pairing.

An inner product on $V$ is a positive-definite bilinear form that
induces norms, distances, and angular quantities. For example,
\begin{equation}
    \|v\| = \sqrt{\langle v, v \rangle},
\end{equation}
or
\begin{equation}
    \cos(u,v) = \frac{\langle u, v \rangle}{\|u\|\|v\|}.
\end{equation}
Such quantities depend on the chosen inner product and are therefore
not intrinsic to a vector space alone.

\section{Proof of Proposition~\ref{prop:primal-dual}(i)}

Let $f : V^{*} \to V$ satisfy $f(A^{-\top}w) = A f(w)$ for all $A \in \mathrm{GL}(V)$ and all
$w \in V^{*}$. Fix $w \neq 0$ and let
$H_w = \{A \in \mathrm{GL}(V) : A^{-\top}w = w\}$ be its stabilizer under the dual action, equivalently
$\{A : w \circ A = w\}$. Equivariance gives $f(w) = A f(w)$ for every $A \in H_w$, so $f(w)$
lies in the subspace of $V$ fixed pointwise by $H_w$.

Write $K = \ker w$, a hyperplane, and pick $u \notin K$. Every $A$ preserving $w$ acts
arbitrarily on $K$ and fixes $u$ modulo $K$, so $H_w$ contains all maps of the form
$u \mapsto u + \kappa$, $\left.A\right|_{K} \in \mathrm{GL}(K)$ with $\kappa \in K$ arbitrary. Let
$v \in V$ be fixed by all of $H_w$. Writing $v = \alpha u + \kappa_0$ with $\kappa_0 \in K$
and applying $u \mapsto u + \kappa$ gives $\alpha\kappa = 0$ for all $\kappa \in K$, hence
$\alpha = 0$ whenever $\dim V \ge 2$. Then $v = \kappa_0 \in K$ is fixed by all of
$\mathrm{GL}(K)$, which forces $\kappa_0 = 0$ since $\dim K \ge 1$ and $\mathrm{GL}(K)$ acts transitively on
$K \setminus \{0\}$. Therefore $f(w) = 0$ for every $w \neq 0$, and $f(0) = 0$ follows from
equivariance under $A = cI$ with $c \neq 1$. Hence $f \equiv 0$.

The hypothesis $\dim V \ge 2$ cannot be dropped. On $V = \mathbb{R}$ the dual action is
$w \mapsto w/a$ and the map $f(w) = c/w$ satisfies $f(w/a) = a f(w)$ for every $a \neq 0$, so
nonzero equivariant maps exist in dimension one.

\section{Architecture-Induced Representation Symmetries}
\label{app:arch}

\subsection{The rotary commutant}

Let $d_{\mathrm{head}} = 2n$ and let RoPE act at position $t$ by
$R(t) = \bigoplus_{j=1}^{n} R_{2}(t\theta_{j})$, where $R_{2}(\alpha)$ is the planar rotation
by $\alpha$ and $\theta_{1},\dots,\theta_{n}$ are the rotary frequencies. The attention logit
between a query at position $t$ and a key at position $s$ is
\begin{equation}
\langle R(t) W_Q x, R(s) W_K y \rangle
= x^{\top} W_Q^{\top} R(s-t) W_K \, y ,
\end{equation}
using $R(t)^{\top}R(s) = R(s-t)$.Writing $\tau=s-t$, a reparameterization $W_Q \mapsto \Lambda W_Q$,
$W_K \mapsto \Lambda' W_K$ preserves every logit if and only if
\begin{equation}
\Lambda^{\top} R(\tau) \Lambda' = R(\tau)
\qquad \text{for all relative positions } \tau .
\label{eq:rope-condition}
\end{equation}
Taking $\tau = 0$ gives $\Lambda' = \Lambda^{-\top}$, recovering the gauge of
Corollary~\ref{cor:reading-point}. Substituting back, Equation~\ref{eq:rope-condition}
becomes $R(\tau)\Lambda^{-\top} = \Lambda^{-\top}R(\tau)$ for all $\tau$: the admissible
$\Lambda^{-\top}$ are exactly the elements of the commutant of $\{R(\tau)\}_{\tau}$ in
$\mathrm{GL}(2n)$.

Identify $\mathbb{R}^{2n} \cong \mathbb{C}^{n}$ by pairing the two coordinates of each rotary
plane, under which $R(\tau)$ becomes multiplication by
$\mathrm{diag}(e^{i\tau\theta_{1}},\dots,e^{i\tau\theta_{n}})$. When the frequencies are
pairwise distinct and the relative positions $\tau$ range over enough values to separate
them, the $\mathbb{R}$-algebra generated by $\{R(\tau)\}_{\tau}$ is the full diagonal algebra
$\mathbb{C}^{n}$, whose commutant in $\mathrm{End}_{\mathbb{R}}(\mathbb{R}^{2n})$ is again
$\mathbb{C}^{n}$. Its invertible elements are
\begin{equation}
\Lambda^{-\top} = \bigoplus_{j=1}^{n} s_{j} R_{2}(\alpha_{j}),
\qquad s_{j} > 0, \ \alpha_{j} \in [0,2\pi),
\label{eq:rope-image}
\end{equation}
so $\operatorname{Im}\rho$ at a key site is the group of independently scaled rotations of the
rotary planes, of real dimension $2n = d_{\mathrm{head}}$. Degenerate frequency sets enlarge
the commutant: if $m$ frequencies coincide, the corresponding block is $\mathrm{GL}_{m}(\mathbb{C})$
rather than $\mathbb{C}^{\times}$, so Equation~\ref{eq:rope-image} is the smallest image the
architecture realizes.

\subsection{Consequences at a key site}

Equation~\ref{eq:rope-image} is a proper subgroup of $\mathrm{GL}(d_{\mathrm{head}})$, so the argument
of Corollary~\ref{cor:reading-point} does not apply directly. It nonetheless excludes the
metric predicates in question, because the scale factors $s_{j}$ vary independently across
planes. Write $k = (k^{(1)},\dots,k^{(n)})$ in rotary-plane blocks. Under
Equation~\ref{eq:rope-image},
\begin{equation}
\|k\|^{2} \mapsto \sum_{j} s_{j}^{2} \|k^{(j)}\|^{2},
\end{equation}
which equals $\|k\|^{2}$ for all $k$ only when every $s_{j} = 1$. Norms are therefore not
invariant, and neither are cosines: choosing $s_{1} \neq s_{2}$ and $s_{j} = 1$ otherwise
changes the relative weight of the first two planes in $\langle k_{1},k_{2}\rangle$ while
changing the norms differently, so $\cos(k_{1},k_{2})$ is not preserved. The same
anisotropy moves the eigenvalues of $KK^{\top}$ non-uniformly, so PCA directions,
$\log\det(KK^{\top})$, and any spectral quantity not expressible through ratios fixed by the
per-plane scaling are likewise gauge dependent. Only a uniform scaling
$s_{1} = \cdots = s_{n}$ would restrict the action to $G_{\mathrm{sim}}$ and rescue the angular
quantities, and the architecture does not impose it.

By contrast, the criterion of Section~\ref{sec:invariant-alternative} survives, since
Equation~\ref{eq:rope-image} is a subgroup of the gauge under which $\Omega$ and the key
differences were shown to transform oppositely. The RoPE restriction therefore shrinks the
architectural image without restoring any of the quantities Corollary~\ref{cor:reading-point}
excludes.

\subsection{What an invariant version looks like}
\label{sec:invariant-alternative}

The framework does more than reject quantities. Because the QK gauge preserves realised attention logits $q^\top k$ exactly, any predicate expressible through those logits is admissible at these sites. Without RoPE, the corresponding parameter-level object is $W_Q^\top W_K$; with RoPE, it is the family $W_Q^\top R(\tau)W_K$ indexed by relative position. This is enough to restate the eviction criterion invariantly. What KeyDiff needs is a notion of when two realised keys are interchangeable in their effect on attention, and that effect is mediated through $q^\top(k_1-k_2)$. Define the second moment of the realised queries by
\begin{equation}
\Omega = \mathbb{E}\!\left[q q^{\top}\right],\qquad d(k_1,k_2)^{2} = (k_1-k_2)^{\top}\,\Omega\,(k_1-k_2).
\end{equation}
In the no-RoPE case, $\Omega=W_Q\Sigma W_Q^\top$, where $\Sigma$ is the second moment of the residual-stream inputs on the query side.
Under the induced gauge $q\mapsto\Lambda q$ and $k\mapsto\Lambda^{-\top}k$, we have $\Omega\mapsto\Lambda\Omega\Lambda^\top$ and $k_1-k_2\mapsto\Lambda^{-\top}(k_1-k_2)$, so the two transformations cancel and $d$ is unchanged. The quantity it measures is also the right one: $d(k_1,k_2)^{2}$ is the mean squared change in attention logit incurred by substituting one key for the other. When $\Omega \propto I$ the criterion reduces to Euclidean distance in key space, which orders pairs as cosine does when key norms are equal, so the original rule is recovered as the special case in which the realized query distribution is isotropic in the gauge that happens to have been sampled. The distinction is not merely formal: gauge-related parameter settings compute the same model function while producing different eviction sets under the cosine rule and identical eviction sets under $d$.


\subsection{Head-space quantities}
\label{app:head-space}

The same architectural gauge constrains geometric comparisons between attention heads. \citet{fokouuavs2026multi} compute cross-Gram matrices between key-projection matrices and interpret their singular values in terms of principal angles between head subspaces. If $G_{hh'}$ denotes such a cross-Gram matrix, independent invertible changes of basis within the two heads give
\begin{equation}
G_{hh'} \mapsto B_h^\top G_{hh'} B_{h'}.
\end{equation}
Its singular values are therefore not invariant under general invertible $B_h$ and $B_{h'}$ unless the spanning matrices have first been orthonormalised. This does not imply that the underlying subspaces are gauge dependent: when represented in residual-stream coordinates, those subspaces are unchanged by invertible changes of basis within the heads.

\citet{ge2026prism} compare leading left singular vectors of different heads by cosine similarity and calibrate the observed statistic against uniformly sampled unit vectors in a fixed Euclidean $\mathbb{R}^{d_{\mathrm{head}}}$. Both constructions depend on the chosen head-space geometry. An anisotropic change of head coordinates changes the cosine statistic, while the uniform distribution on the Euclidean unit sphere is itself not preserved by such a transformation.

\end{document}